\documentclass[lettersize,journal]{IEEEtran}

\usepackage{graphicx}
\usepackage{stfloats}
\usepackage{amsmath}
\usepackage{booktabs}
\usepackage{amssymb}
\usepackage{hyperref}
\usepackage{threeparttable}  
\usepackage{tabularx}
\usepackage{cite}
\usepackage{textcomp}
\usepackage{stfloats}
\usepackage{url}
\usepackage{verbatim}
\usepackage{amsfonts}

\begin{document}
\title{Beyond Segmentation: An Oil Spill Change Detection Framework Using Synthetic SAR Imagery}
\author{Chenyang Lai, Shuaiyu Chen, Tianjin Huang, Siyang Song, Guangliang Cheng, Chunbo Luo, Zeyu Fu
\thanks{Chenyang Lai, Shuaiyu Chen, Tianjin Huang, Siyang Song, Chunbo Luo and Zeyu Fu are with the Department of Computer Science, University of Exeter, Exeter,
UK. Emails: sc1321@exeter.ac.uk, t.huang2@exeter.ac.uk, s.song@exeter.ac.uk, c.luo@exeter.ac.uk, z.fu@exeter.ac.uk}
\thanks{Guangliang Cheng is with the Department of Computer Science, University of Liverpool, Liverpool,
UK. Email: guangliang.cheng@liverpool.ac.uk}
}
\markboth{}%
{Shell \MakeLowercase{\textit{et al.}}: A Sample Article Using IEEEtran.cls for IEEE Journals}
\maketitle    
\begin{abstract}
Marine oil spills are urgent environmental hazards that demand rapid and reliable detection to minimise ecological and economic damage. While Synthetic Aperture Radar (SAR) imagery has become a key tool for large-scale oil spill monitoring, most existing detection methods rely on deep learning-based segmentation applied to single SAR images. These static approaches struggle to distinguish true oil spills from visually similar oceanic features (e.g., biogenic slicks or low-wind zones), leading to high false positive rates and limited generalizability, especially under data-scarce conditions.
To overcome these limitations, we introduce Oil Spill Change Detection (OSCD), a new bi-temporal task that focuses on identifying changes between pre- and post-spill SAR images. As real co-registered pre-spill imagery is not always available, we propose the Temporal-Aware Hybrid Inpainting (TAHI) framework, which generates synthetic pre-spill images from post-spill SAR data. TAHI integrates two key components: High-Fidelity Hybrid Inpainting for oil-free reconstruction, and Temporal Realism Enhancement for radiometric and sea-state consistency. Using TAHI, we construct the first OSCD dataset and benchmark several state-of-the-art change detection models. Results show that OSCD significantly reduces false positives and improves detection accuracy compared to conventional segmentation, demonstrating the value of temporally-aware methods for reliable, scalable oil spill monitoring in real-world scenarios.
\begin{IEEEkeywords}
Oil spill change detection, Synthetic Aperture Radar, Remote Sensing
\end{IEEEkeywords}

\end{abstract}

\section{Introduction}
Marine oil spills pose substantial threats to marine ecosystems, coastal economies, and public health, necessitating rapid and precise detection methods. Space-borne Synthetic Aperture Radar (SAR), capable of all-weather and day-night imaging, has become a cornerstone technology for large-scale oil spill monitoring~\cite{rs12203338,Amri_2021,Brekke2005}. Effective SAR-based detection significantly enhances containment and mitigation efforts, minimising environmental and economic damages.
\begin{figure}[t]
    \centering
    \includegraphics[width=\linewidth]{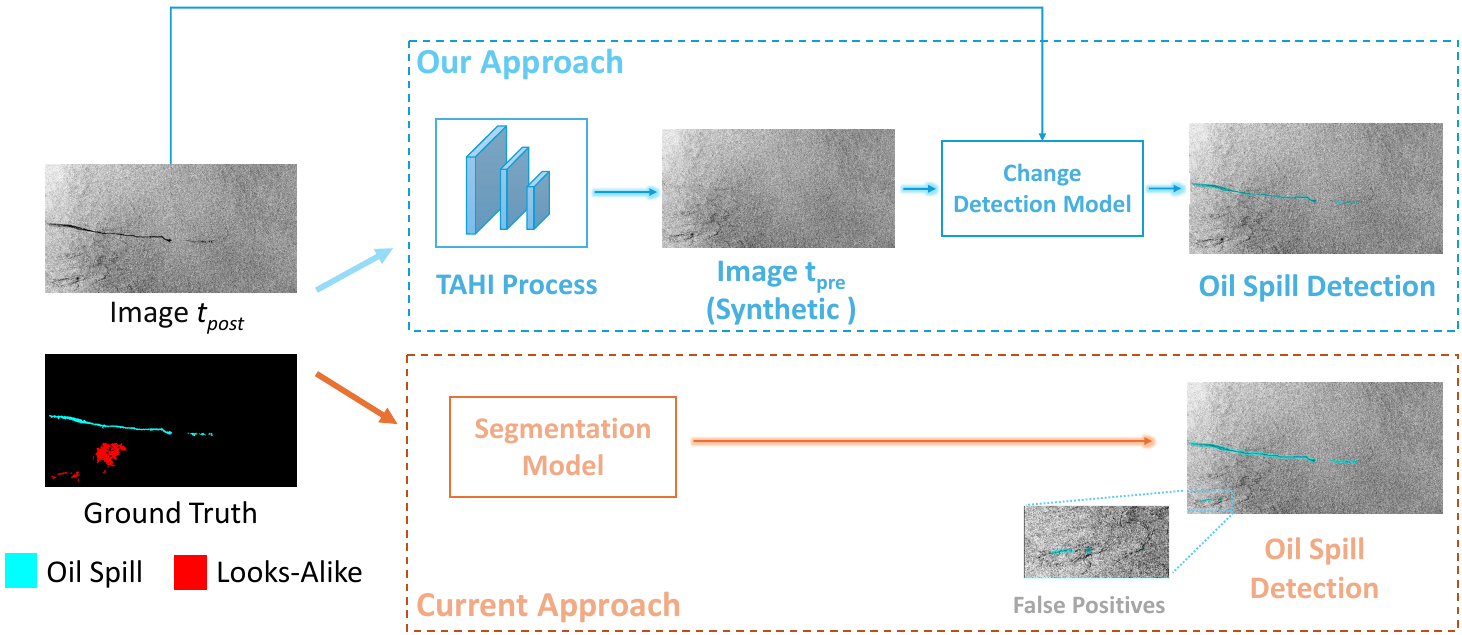}
    \caption{Comparison between our proposed approach and current segmentation-based methods. Our temporally aware change detection approach effectively isolates new oil spills, reducing false positives caused by visually similar look-alikes.}
    \label{fig:approach_comparison}
\end{figure}

Current automated oil spill detection pipelines predominantly rely on deep learning–based segmentation models applied to single post-event SAR images~\cite{Alhashmi2025DeepLab,Chen2022,Krestenitis2019,TransOilSeg10909301,OSDMamba2025,LI2023113872,Guo2024FewShot,10485498}. These data-driven approaches can capture richer spatial patterns than classical thresholding and texture-based methods and have shown promising performance on curated benchmarks. At the same time, operational services still depend heavily on expert visual interpretation, ancillary met-ocean information, and physics-based screening tools to contextualise SAR observations~\cite{rs12203338,Brekke2005}. From a modelling perspective, most learning-based methods still treat each SAR acquisition independently and do not explicitly exploit the short-term temporal context that is routinely inspected qualitatively by human analysts. As a result, single-image segmenters often struggle to differentiate true oil spills from visually similar phenomena, such as biogenic slicks, low-wind zones, internal waves, or calm sea states, leading to frequent false positives in dynamic marine environments (see Fig.~\ref{fig:approach_comparison}). The problem is further compounded by the scarcity of annotated oil spill data and the severe class imbalance between spill and non-spill pixels, both of which limit model generalisability~\cite{Mahmoudi2022,OSDMamba2025}. In this work, we therefore view deep learning as an emerging, complementary tool for exploiting temporal information, rather than as a replacement for existing physics-informed practice.

To address these challenges, we advocate a shift toward temporally aware detection via bi-temporal change detection, introducing a new task: oil spill change detection (OSCD), as illustrated in Fig.~\ref{fig:approach_comparison}. Rather than attempting to infer “spillness’’ from a single snapshot, OSCD explicitly analyses differences between pre- and post-spill SAR images. This formulation improves upon traditional segmentation paradigms by focusing on temporal differences rather than static appearance, enabling the model to more effectively suppress quasi-persistent background patterns and look-alike phenomena within the short observation window while emphasising genuinely new spill-related structures. Here, “quasi-persistent’’ refers to features that evolve more slowly than the time interval between the pre- and post-event acquisitions (e.g., shipping lanes or infrastructure footprints), rather than implying that the ocean surface is strictly static. Unlike temporal smoothing or recurrent segmentation, OSCD explicitly reframes oil spill detection as a change localization problem, where persistence itself becomes a discriminative cue rather than noise.

Implementing OSCD in practice is, however, challenging due to the limited availability of co-registered pre-spill imagery. In real operational settings, suitable pre-event SAR acquisitions with comparable incidence angles, sea states, and acquisition geometries are rarely available, as ocean dynamics, satellite revisit intervals, and mission-dependent imaging modes jointly complicate the acquisition of temporally aligned bi-temporal pairs. These constraints severely limit the direct application of standard bi-temporal CD techniques in maritime spill monitoring.

To overcome this limitation, we propose the Temporal-Aware Hybrid Inpainting (TAHI) framework, which synthesises pre-spill SAR images from annotated post-spill data. TAHI consists of two key components: (1) High-Fidelity Hybrid Inpainting (HFHI), which removes oil-contaminated regions using SAR-aware hybrid patch-based inpainting and reconstructs the underlying sea-surface texture; and (2) Temporal Realism Enhancement (TRE), which refines the inpainted result by aligning local radiometric statistics and injecting simulated environmental dynamics, such as subtle wave patterns, speckle fluctuations, and low-frequency sea-state drift, to emulate realistic inter-pass variability. By pairing each post-spill image with its corresponding synthetic pre-spill counterpart, TAHI enables the construction of a bi-temporal OSCD dataset suitable for supervised CD training and systematic benchmarking.

Crucially, our framework is designed as an offline data generation pipeline that can be integrated into realistic multi-temporal surveillance workflows. In an operational scenario, agencies maintain short temporal stacks of SAR imagery over areas of interest. CD models trained on our OSCD dataset can then be deployed as temporal filters: candidate spills detected in a newly acquired scene are cross-checked against recent pre-event acquisitions, allowing persistent look-alike phenomena and sensor artefacts to be systematically down-weighted while emphasising genuinely new anomalies. This multi-temporal perspective naturally complements existing single-image detectors and expert analysis.
Our key contributions are as follows. 
\begin{itemize}
    \item We introduce the OSCD task and release the first bi-temporal change detection dataset specifically designed for oil spill monitoring. To the best of our knowledge, this is among the first studies to systematically reframe SAR-based oil spill detection as a supervised deep learning–based change-detection problem.
    \item We propose the TAHI framework, which synthesises realistic pre-spill SAR images from post-spill observations by combining HFHI with TRE, enabling high-quality reconstruction of oil-free scenes with plausible sea-surface dynamics.
    \item 
    Experimental results show that our approach enables robust training of advanced CD models, significantly improving oil spill detection performance compared to conventional segmentation methods and providing a scalable foundation for integrating temporally aware oil-spill monitoring into future multi-temporal environmental surveillance systems. Importantly, the synthetic pre-spill images are not intended to approximate exact historical observations, but to provide statistically consistent temporal contrast for supervised CD training.
\end{itemize}


\section{Related Work}
\subsection{SAR-based Oil Spill Detection}
Early SAR-based oil spill detection methods relied on hand-crafted heuristics and classical image processing techniques, such as intensity thresholding, texture descriptors, and basic statistical classifiers, to identify dark formations presumed to indicate oil slicks against the surrounding sea surface~\cite{rs12203338,Brekke2005}. Although these approaches are simple and computationally efficient, they exhibit limited robustness under complex sea states and diverse look-alike phenomena and often require substantial expert intervention, restricting their operational scalability. Comprehensive reviews~\cite{rs12203338} highlight that purely rule-based systems struggle to generalise across sensors, acquisition modes, and environmental conditions.

The advent of deep learning significantly enhanced segmentation capabilities for SAR-based oil spill detection. Del Frate et al.~\cite{DelFrate2000} first validated neural networks on ERS-1 imagery, and subsequent work leveraged larger Sentinel-1 datasets with fully convolutional networks and attention U-Nets for further gains~\cite{Amri_2021,Chen2022,Krestenitis2019}. More recent models explicitly address data limitations and look-alike confusion: YOLO-based frameworks combined with the Segment Anything Model~\cite{Kirillov_SAM} (YOLOv8-SAM\cite{10485498}) provide real-time object-level detection with strong localisation speed, while transformer-driven architectures such as TransOilSeg~\cite{TransOilSeg10909301} capture long-range spatial dependencies for finer boundary delineation. DeepLabv3+-based approaches that model speckle statistics~\cite{Alhashmi2025DeepLab,LI2023113872} and the state-space–driven OSDMamba architecture~\cite{OSDMamba2025} further improve pixel-level accuracy, particularly under class-imbalanced conditions~\cite{Mahmoudi2022,OSDMamba2025}. Complementary to these architectures, Chen~et~al.\ propose a self-supervised boundary-aware block for marine pollution segmentation that improves boundary localisation under weak and ambiguous edges while remaining independent of manual boundary annotations, and can be seamlessly integrated into diverse backbones~\cite{chen2025enhancing}.

Beyond purely data-driven segmentation, a substantial body of work has investigated physically grounded approaches for SAR-based oil monitoring. These methods exploit the radar backscattering mechanism, wind and current fields, and, where available, polarimetric information. For example, Chen and Wang~\cite{Chen2022} incorporate polarimetric channels and wind speed into an attention U-Net architecture to better discriminate mineral oil from biogenic slicks and low-wind areas, while Brekke and Solberg~\cite{Brekke2005} and Al-Ruzouq et al.~\cite{rs12203338} summarise how incidence-angle normalisation, met-ocean data, and expert knowledge are integrated into operational services. Our OSCD formulation is complementary to these physics-aware approaches: it assumes only single-polarisation intensity data, and the synthetic pre/post-spill pairs generated by TAHI are intended to provide an additional temporal cue that could be combined with such model-based pre-screening tools in future surveillance chains.

\subsection{Change Detection in Remote Sensing}
Change detection (CD), which analyses differences between images acquired at different times, has emerged as a core paradigm for remote sensing applications and provides a natural alternative to static image segmentation~\cite{Bai2023}. Deep learning approaches to CD typically ingest bi-temporal input pairs to identify changes by comparing pre- and post-event images~\cite{Bai2023}. Recent advances employ deep convolutional networks, transformer-based architectures, and enhanced interaction/fusion designs to boost both accuracy and robustness~\cite{Bai2023,zheng2023changeeverywheresingletemporalsupervised,ma2024ddlnetboostingremotesensing}. Representative bi-temporal architectures include fully convolutional Siamese networks with early feature fusion~\cite{CDfc}, star-shaped decoders that exploit multi-scale context~\cite{zheng2023changeeverywheresingletemporalsupervised}, and cross-interaction modules that explicitly model spatio-temporal dependencies~\cite{ma2024ddlnetboostingremotesensing}. Despite this progress, most CD techniques have not yet been effectively applied to oil spill detection in practice, partly because well-aligned pre-event SAR imagery with comparable imaging conditions is difficult to obtain due to satellite revisit intervals, orbit constraints, and the stochastic timing of spills.

\section{Methodology}\label{sec:method}
\subsection{Problem Formulation}
We formulate the overall task as Oil Spill Change Detection, as shown in Fig. \ref{fig:approach_comparison}, which aims to identify spill-specific changes between bi-temporal SAR observations while suppressing persistent background structures and look-alike phenomena. This task naturally aligns with the broader paradigm of CD, which compares pre- and post-event imagery to localize changes over time. However, in real-world maritime monitoring scenarios, co-registered pre-spill SAR images with similar imaging conditions are rarely available due to limited satellite revisit rates and unpredictable spill timing. This severely limits the direct application of standard bi-temporal CD techniques.

To overcome this challenge, we propose generating synthetic pre-spill images.  Specifically, given a post-spill SAR image $I_t$ and its corresponding oil mask $M$, we synthesise a pre-spill image $\hat{I}_{t-1}$ that approximates the uncontaminated sea surface prior to the spill. This resulting image pair $(\hat{I}_{t-1}, I_t)$ is then fed into a bi-temporal CD model to produce a binary change map $C$, highlighting regions affected by the spill.
This generation is achieved using our newly designed Temporal-Aware Hybrid Inpainting (TAHI) framework.
In the following subsection, we describe the TAHI framework and its components.
\subsection{Temporal-Aware Hybrid Inpainting}
\begin{figure}[t]
    \centering
    \includegraphics[width=1\linewidth]{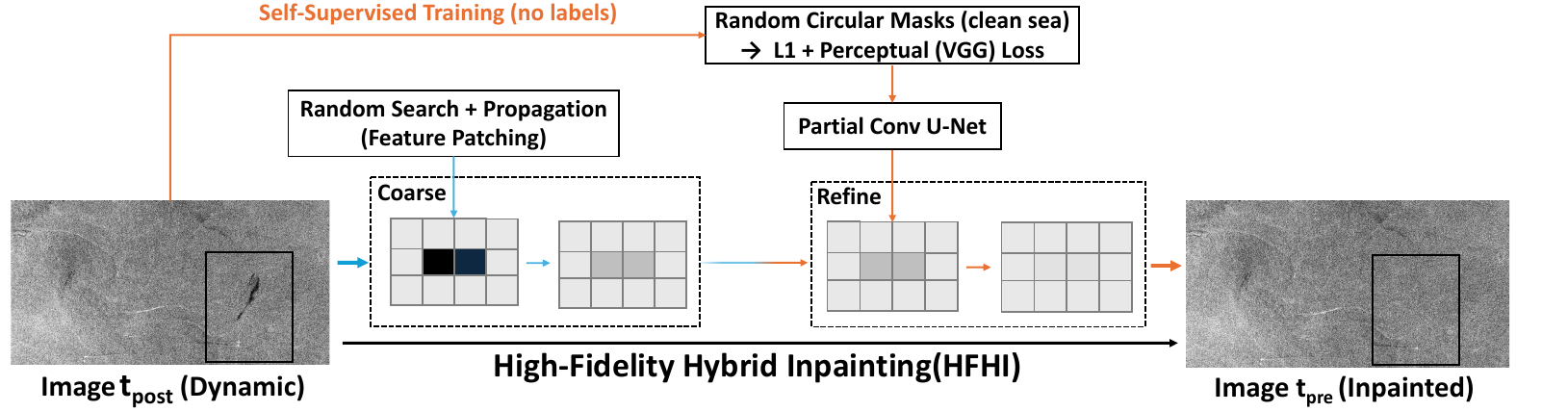}
\caption{TAHI Component 1: High-Fidelity Hybrid Inpainting (HFHI) restores spatial continuity to generate a pre-event estimate.}
    \label{fig:tahis2}
\end{figure}
\subsubsection{High-Fidelity Hybrid Inpainting}
In the first component of our framework, the High-Fidelity Hybrid Inpainting (HFHI) applies the PatchMatch algorithm~\cite{Barnes2009} to the preprocessed post-event image \textit{Image}~$t_{\text{post}}$~(Dynamic) to reconstruct regions affected by the oil spill (label = 1) and areas vacated due to vessel removal or displacement. These regions are jointly masked and treated as missing during the inpainting process (see \ref{fig:tahis2}). Let \( I: \mathbb{Z}^2 \rightarrow \mathbb{R} \) denote the grayscale SAR image, and \( M: \mathbb{Z}^2 \rightarrow \{0,1\} \) the binary inpainting mask, where \( M(x,y) = 1 \) indicates a missing pixel to be inpainted. For each masked pixel \( (x,y) \) with \( M(x,y) = 1 \), the goal is to find a best-matching location \( (x', y') \) in the known region \( \mathcal{K} = \{(u,v) \in \mathbb{Z}^2 \mid M(u,v) = 0\} \) such that the local patch centred at \( (x', y') \) is most similar in appearance to that centred at \( (x,y) \). We define a square patch \( P_{(x,y)} \in \mathbb{R}^{(2r+1)\times(2r+1)} \) of radius \( r \) around pixel \( (x,y) \), and obtain the nearest neighbour by
\begin{equation}
(x', y') =
\operatorname*{arg\,min}_{(u,v)\in\mathcal{K}}
\bigl\| P_{(x,y)} - P_{(u,v)} \bigr\|_2^2 ,
\label{eq:patchmatch_nn}
\end{equation}
where patch similarity is measured using the squared Euclidean distance between pixel intensities. The PatchMatch algorithm~\cite{Barnes2009} efficiently approximates this nearest-neighbour field in \ref{eq:patchmatch_nn}, which maps each masked pixel to its most similar patch in the known (unmasked) region. Although computationally efficient, PatchMatch may introduce visible seams or repetitive artifacts when the missing region is large, particularly in homogeneous ocean surfaces common in SAR imagery. PatchMatch excels at preserving global SAR texture statistics but struggles with large homogeneous regions. While partial-convolution networks provide local regularization but risk over-smoothing; HFHI combines their complementary strengths~\cite{deledalle2014nonlocal,guillemot2014inpainting,yu2018contextual,Liu_2018_ECCV}.

To mitigate this, we further refine the inpainting results using a deep learning module based on a U-Net architecture with partial convolutions~\cite{Liu_2018_ECCV}. Unlike standard convolutions, partial convolutions update only valid (non-masked) pixels, thereby preserving local texture and structural continuity. In practice, partial-convolution networks provide local regularization but can risk over-smoothing fine-grained speckle-like patterns; HFHI combines the complementary strengths of PatchMatch and partial convolutions to achieve both global statistical consistency and locally coherent boundaries.

Specifically, multiple random circular masks $\{M_k\}$ of varying sizes and counts are generated over clean regions, and the network is trained by minimizing the following loss:
\begin{equation}
\mathcal{L} = \bigl\| f(I \odot M_k) - I \bigr\|_1
+ \lambda \, \bigl\| \phi\!\bigl(f(I \odot M_k)\bigr) - \phi(I) \bigr\|_1 ,
\label{eq:inpaint_loss}
\end{equation}
where \( f \) is the inpainting network, \( \phi \) represents features extracted from a pre-trained VGG network~\cite{simonyan2015} (used for perceptual loss~\cite{Zhang2016PerceptualLoss}), and \( \lambda \) balances reconstruction and perceptual fidelity. Although VGG features are learned from optical imagery, we employ them only as a weak structural regularizer to suppress patch-wise discontinuities, rather than as a semantic prior.~\cite{dosovitskiy2016deepsim,zhang2018perceptualmetric} The objective in \ref{eq:inpaint_loss} encourages the network to restore artificially masked areas while learning the intrinsic backscatter patterns and intensity distributions characteristic of the ocean surface. All training hyperparameters—including the mask size range, \(\lambda\), and optimizer settings—were selected on a held-out validation subset to minimize boundary artifacts and improve texture smoothness.

To further enhance boundary consistency, the original spill mask is slightly dilated to eliminate subtle oil fringes, ensuring a smooth transition across edges. Once trained, the network is applied to the entire oil-contaminated region in a single forward pass, producing the restored \textit{Image $t_{\text{pre}}$ (Inpainted)} which preserves both fine-scale speckle characteristics and the coherent structure of the sea surface.

\begin{figure}[t]
    \centering
    \includegraphics[width=1\linewidth]{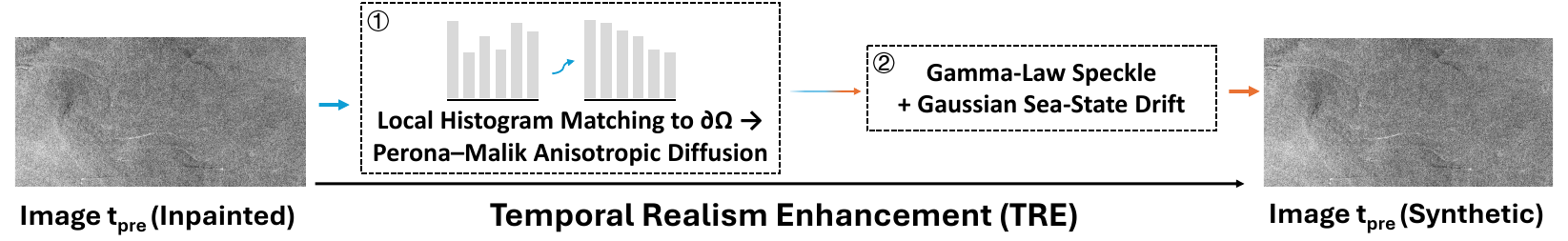}
\caption{TAHI Component 2: Temporal Realism Enhancement (TRE) refines radiometric and textural properties to produce the final reference image.}
    \label{fig:tahis3}
\end{figure}

In the geospatial domain, void filling and inpainting have also been studied for other modalities such as digital surface models. For example, Panangian and Bittner~\cite{Panangian_2025_WACV} propose DFILLED, which repurposes deep anisotropic diffusion models for guided DSM void filling using inpainting masks that mimic natural void patterns. In contrast, TAHI focuses on SAR intensity imagery and oil-contaminated regions, combining exemplar-based PatchMatch and partial-convolution refinement to reconstruct sea-surface backscatter while preserving speckle statistics.

\subsubsection{Temporal Realism Enhancement}
\label{sec:tra}
After applying HFHI, we refine the radiometric consistency and introduce scene dynamics through two sequential steps (see \ref{fig:tahis3}).
First, to ensure that the global brightness and contrast of the inpainted oil spill region align with the surrounding sea clutter, we perform empirical cumulative histogram matching.
Specifically, the intensity distribution within the masked spill region $\Omega$ is aligned to its eight-connected boundary $\partial\Omega$ as follows:
\begin{equation}
I_{\text{eq}}(x,y) =
H_{\partial\Omega}^{-1}\!\left(
F_{\Omega}\bigl(t_{\text{pre}}^{\text{(Inpainted)}}(x,y)\bigr)
\right),
\label{eq:hist_match}
\end{equation}
where $F_\Omega$ denotes the empirical CDF within the inpainted mask, and $H_{\partial\Omega}^{-1}$ is the inverse cumulative distribution functions (CDFs) computed from the neighbouring pixels along $\partial\Omega$.
The equalized image $I_{\text{eq}}$ in \ref{eq:hist_match} enforces local radiometric consistency between the reconstructed spill and its surroundings.

To mitigate edge discontinuities and enhance radiometric smoothness, we then apply Perona–Malik anisotropic diffusion~\cite{56205} for 20 iterations with $\kappa{=}15$, constrained to a 5-pixel band surrounding $\partial\Omega$.
The diffusion iterations and $\kappa$ were tuned to suppress boundary seams while preserving texture, whereas the speckle looks ($L{=}4$) and drift strength ($\alpha{=}0.05$) were set to match background ENL/CNR statistics on the validation set.
This diffusion process reduces gradient seams between the inpainted and original regions, resulting in a seamless transition.
We further apply two statistically grounded perturbations:
(i) multiplicative speckle noise, modeled as $\eta(x,y)\!\sim\!\Gamma\!\bigl(L,1/L\bigr)$ with $L{=}4$ looks, introducing stochastic roughness; and  
(ii) low-frequency radiometric drift, expressed as a multiplicative term $1+\alpha G(x,y)$, where $G$ is a zero-mean Gaussian random field smoothed by a $51{\times}51$ box filter and $\alpha{=}0.05$, representing gradual sea state changes such as wind-induced streaks.
The final synthetic pre-event view is then given by
\begin{equation}
\widehat I_{t-1}(x,y)
= I_{\text{eq}}(x,y)\,\eta(x,y)\,\bigl[1+\alpha\,G(x,y)\bigr],
\label{eq:synthetic_pre}
\end{equation}
where $\widehat I_{t-1}$ denotes the generated pre-spill image and $I_{t}$ is the original post-spill observation.
These perturbations ensure that while the global SAR statistics are preserved, the only spatially coherent and systematic change between the synthetic pre-spill image $\widehat I_{t-1}$ and the original post-spill image $I_{t}$ is the presence of the oil slick.

\section{Dataset Construction and Ablation Study}

\subsection{OSCD Dataset}
Our dataset is constructed based on the M4D dataset~\cite{Krestenitis2019}, which contains Sentinel-1 SAR VV-polarised intensity images and corresponding masks annotated with various maritime features. We select all samples with confirmed oil spills, as indicated in the metadata and masks, to serve as post-event (polluted) inputs in our framework. The original masks are used as ground truth, where label 1 indicates spill regions and labels 0, 2, 3, and 4 are treated as background or distractors. Consistent with the design of M4D, our study focuses on surface slicks that manifest as extended dark formations in VV-polarised backscatter; very thin sheens or strongly fragmented slicks that remain close to the sensor noise floor are largely absent from the annotations and therefore remain outside the scope of the present work. To account for dynamic vessel behaviour, the preprocessing stage perturbs vessel positions in the post-event SAR images, simulating plausible pre-spill vessel motion or disappearance.

The resulting OSCD dataset comprises 879 bi-temporal SAR image pairs, each coupled with a verified binary change mask. The dataset is split into 791 training samples and 88 test samples, ensuring that each oil spill scenario appears in only one subset to avoid data leakage. To better understand the dataset composition and the inherent class imbalance, we provide pixel-level statistics of the post-spill masks in Table~\ref{tab:pixel_stats}, where oil pixels account for roughly 1\% of all pixels in both training and test subsets.

\begin{table}[t]
    \centering
    \caption{Pixel-level statistics of the post-spill masks in training and test subsets.}
    \label{tab:pixel_stats}
    \begin{tabular}{lccc}
        \toprule
        Subset & Image pairs &  Oil Pixels  &  Oil Pixels Ratio \\
        \midrule
        Training  & 791 & 8,164,677 & 1.02\% \\
        Test      & 88  &   955,554 & 1.07\% \\
        \bottomrule
    \end{tabular}
\end{table}

Beyond serving as a dataset for training our own CD models, OSCD is designed as a reusable benchmark for the wider community. The synthetic pre-spill images generated by TAHI are only required during dataset construction; at deployment time, practitioners are expected to pair newly acquired post-event SAR images with archived pre-event acquisitions collected along the same shipping lanes or around the same offshore infrastructure. A CD model trained on OSCD can then analyse these real bi-temporal stacks and highlight genuinely new dark formations while suppressing persistent clutter and look-alikes. In this way, OSCD connects curated historical annotations, such as those provided by M4D~\cite{Krestenitis2019}, with future multi-temporal monitoring workflows, and it can naturally be extended to other rare maritime events whenever suitable post-event masks are available for offline dataset generation. Notably, the original spill masks are used only during offline dataset construction and are never exposed to CD models at inference time.

\subsection{Restoration Ablation Study}
We next conduct an ablation study to quantify the contribution of each restoration component in the TAHI framework using the following image-quality metrics: Equivalent Number of Looks (ENL), Contrast-to-Noise Ratio (CNR), Integrated and Peak Side-Lobe Ratios (ISLR/PSLR), and a task-driven Dice score obtained from a U-Net++ oil-spill segmenter. The preprocessing stage, which accounts for vessel dynamics, is fixed across all experiments and does not contribute to this ablation. ENL measures noise smoothness over homogeneous sea regions, $\mathrm{ENL}=\mu^2/\sigma^2$~\cite{Li17012024}. CNR, computed between the top and bottom 5\% intensity percentiles, reflects global contrast spread~\cite{Gong15CNR}. ISLR and PSLR quantify residual side-lobe artefacts around the strongest reflector~\cite{Das2010}. The Dice score uses the inpainted mask as pseudo-ground truth; near-zero values indicate that the oil-spill class is no longer detectable.

In qualitative terms, higher ENL corresponds to smoother speckle over homogeneous sea regions, indicating that the inpainted area is statistically consistent with its surroundings. Lower CNR suggests that the restored patch is radiometrically closer to neighbouring clutter, while reduced ISLR and PSLR values imply limited side-lobe leakage around strong reflectors~\cite{Li17012024,Gong15CNR,Das2010}. A near-zero Dice score, computed by running a U-Net++\cite{zhou2019unet++} spill segmenter on the inpainted region and using the inpainting mask as pseudo-ground truth, confirms that residual oil signatures have been effectively removed and are no longer detectable.

\begin{table}[t]
\centering
\caption{Ablation study of restoration components on spill-containing test samples. HFHI = PatchMatch + U\mbox{-}Net; TRE = histogram matching + light anisotropic diffusion (+ optional Gamma-law speckle and Gaussian sea-state drift).}
\label{tab:eval}
\setlength{\tabcolsep}{3.2pt}%
\renewcommand{\arraystretch}{1.05}%
\resizebox{\columnwidth}{!}{%
\begin{tabular}{@{}lccccc@{}}
\toprule
Method & ENL (↑) & CNR (↓) & Dice (↓) & ISLR (dB) ($\approx$) & PSLR (dB) ($\approx$) \\
\midrule
Post-Spill (Original)   & 10.93 & 10.52 & 0.495 & 8.29 & 0.295 \\
+HFHI (PM only)         & 11.11 & 10.14 & 0.106 & 8.33 & 0.316 \\
+HFHI (Full)            & 11.98 &  8.27 & 0.011 & 8.32 & 0.299 \\
+HFHI (Full) + TRE      & \textbf{12.69} & \textbf{7.91} & \textbf{0.003} & 8.29 & 0.296 \\
\bottomrule
\end{tabular}%
}
\end{table}

\begin{figure}[t]
    \centering
    \includegraphics[width=\linewidth]{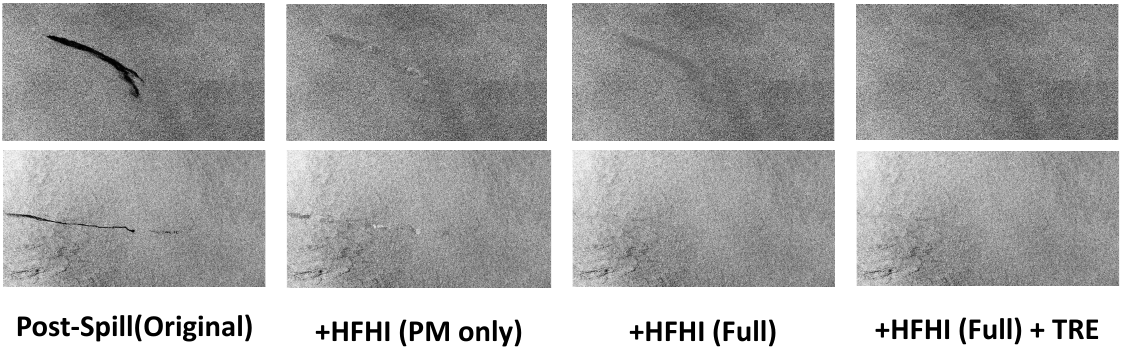}
    \caption{Visual comparison of ablation results. From left to right: Post-Spill (Original), +HFHI (PM only), +HFHI (Full), and +HFHI (Full) + TRE. The proposed pipeline progressively enhances restoration quality, effectively removing oil slicks while preserving sea-surface texture.}
    \label{fig:vis}
\end{figure}

As shown in Fig.~\ref{fig:vis}, the single-module variant (+HFHI~(PM only)) introduces some high-brightness artefacts, whereas +HFHI~(Full) tends to produce smoother but occasionally oversmoothed regions. By incorporating TRE, the complete TAHI framework achieves a more balanced restoration—recovering realistic sea-surface textures while effectively suppressing residual anomalies. Table~\ref{tab:eval} further confirms these observations: the full configuration yields the highest ENL, lowest CNR, and minimal ISLR/PSLR values, while the near-zero Dice score indicates that residual oil signatures are almost entirely removed. Collectively, these findings demonstrate that the components of TAHI work in a complementary manner, yielding the highest overall restoration fidelity.

\section{OSCD Evaluation}
We benchmark both segmentation and change detection (CD) models on the OSCD dataset. As a non-learning baseline, Diff–Otsu computes absolute differences between SAR pairs, $\lvert I_{t_1} - I_{t_0} \rvert$, and applies Otsu thresholding for binary mask generation \cite{otsu1975threshold,XU2011956}. To contextualise CD performance, we additionally report the oil-specific IoU (OSIoU) achieved by four representative single-frame segmentation baselines on post-spill SAR imagery—YOLOv8-SAM \cite{10485498}, DeepLabv3+ \cite{Krestenitis2019}, TransOilSeg \cite{TransOilSeg10909301}, and OSDMamba \cite{OSDMamba2025}. Importantly, these OSIoU values are directly taken from the corresponding original publications (i.e., not re-implemented or re-trained in our pipeline). As the confidence thresholds, post-processing, and evaluation protocols vary across those works, we limit the comparison to OSIoU and refrain from reporting precision, recall, or F$_1$, which would require re-training and re-evaluation under a unified protocol.

For bi-temporal CD, we conduct independent evaluations of four representative architectures using our OSCD splits: CGNet \cite{cgnet} as a compact encoder–decoder baseline, ChangeStar \cite{zheng2023changeeverywheresingletemporalsupervised} with Siamese ResNet backbones and a star-shaped head, FC-Conc \cite{CDfc} with channel-wise feature concatenation, and DDLNet \cite{ma2024ddlnetboostingremotesensing} with dual-domain interaction modules for enhanced spatio-temporal consistency. Together, these networks span early feature concatenation, Siamese differencing, and explicit cross-domain interaction designs, allowing us to probe which temporal fusion mechanisms are most beneficial under the OSCD setting. All CD models are trained and tested using identical data splits, optimised via binary cross-entropy and soft Dice losses, with shared hyperparameters to ensure fair comparison.

Evaluation follows standard metrics, Precision ($\mathcal{P}$), Recall ($\mathcal{R}$), F\textsubscript{1}, and Intersection-over-Union (IoU), with emphasis on oil-specific IoU (OSIoU). Results are averaged across five runs. 

\begin{table}[t]
  \centering
  \setlength{\tabcolsep}{3pt}%
  \renewcommand{\arraystretch}{1.05}%
  \footnotesize
  \begin{tabular}{@{}lcccc@{}}
    \toprule
    Method & OSIoU (\%)$\uparrow$ & $\mathcal{P}$ (\%)$\uparrow$ & $\mathcal{R}$ (\%)$\uparrow$ & $\mathcal{F}_1$ (\%)$\uparrow$ \\
    \midrule
    Baseline & 23.12 & 31.12 & 47.37 & 37.55 \\
    \midrule
    YOLOv8-SAM~\cite{10485498}          & 41.84 & --    & --    & --    \\
    DeepLabv3+~\cite{Krestenitis2019}   & 53.38 & --    & --    & --    \\
    TransOilSeg~\cite{TransOilSeg10909301} & 62.41 & -- & -- & -- \\
    OSDMamba~\cite{OSDMamba2025}        & 65.59 & --    & --    & --    \\
    \midrule
    CGNet~\cite{cgnet}                  & 43.47 & 71.45 & 91.92 & 59.91 \\
    ChangeStar~\cite{zheng2023changeeverywheresingletemporalsupervised} & 67.78 & 75.14 & 87.35 & 80.70 \\
    FC-Conc~\cite{CDfc}                 & 79.58 & 87.28 & 90.02 & 88.62 \\
    DDLNet~\cite{ma2024ddlnetboostingremotesensing} & 82.94 & 88.37 & 93.14 & 90.67 \\
    \bottomrule
  \end{tabular}
  \caption{Quantitative comparison on oil-spill detection (OS = Oil Spill). All values represent the average over five independent test runs.}
  \label{tab:cd_results}
\end{table}

\begin{figure}[ht]
    \centering
    \includegraphics[width=1\linewidth]{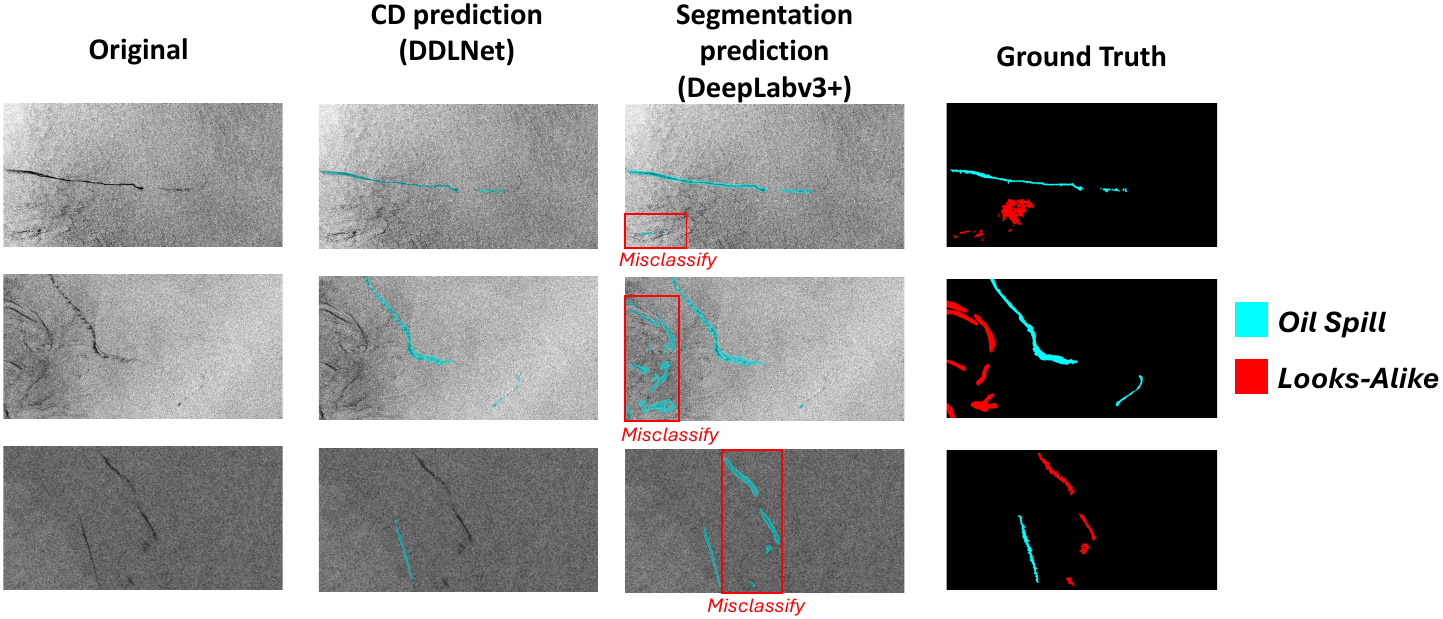}
    \caption{Visual comparison of CD and segmentation outputs on the OSCD dataset. CD better avoids false positives on look-alikes. GT shows oil spills (cyan) and look-alikes (red).}
    \label{fig:cdsgcomp}
\end{figure}

\subsection{Results and Discussion}
As shown in Table~\ref{tab:cd_results}, all four supervised CD models substantially outperform the classical Diff–Otsu baseline. This highlights the effectiveness of the proposed OSCD task, where synthetic bi-temporal supervision (pre-/post-event) provides essential temporal cues absent in single-image approaches, enabling robust CD training even in the absence of real pre-spill observations.

Among the CD models, DDLNet and FC-Conc achieve the best overall performance, balancing high precision and recall. ChangeStar offers a lightweight alternative with competitive accuracy, while CGNet, though architecturally simpler, still delivers significant improvement over the non-learning baseline. These results indicate that, under OSCD, both temporal fusion and architectural design (e.g., cross-feature interaction in DDLNet) are decisive for CD effectiveness. In particular, DDLNet and FC-Conc benefit from stronger cross-temporal interaction: the former couples spatial and spectral cues in a dual-domain fashion, while the latter fuses multi-level features from both timestamps before decoding. The relatively low OSIoU of CGNet, despite using the same training data, suggests that simple encoder–decoder backbones without dedicated temporal fusion are insufficient to fully exploit the bi-temporal supervision provided by OSCD.

For comparison, we also evaluate four segmentation models, YOLOv8-SAM, DeepLabv3+, TransOilSeg, and OSDMamba, that rely solely on post-event imagery. Despite lacking temporal information, their OSIoU scores remain competitive, in some cases even surpassing CGNet. This underscores the value of domain-specific backbones; however, the absence of temporal supervision limits their ability to disambiguate artefacts relative to OSCD-based CD models.

Qualitative results in Fig.~\ref{fig:cdsgcomp} reinforce this observation: CD models trained under the OSCD paradigm consistently avoid false positives on look-alike structures (e.g., biogenic slicks or wind shadows) that often confuse segmentation models. The ground-truth annotations show that CD predictions better match true spill extents, while segmenters tend to over-detect irrelevant features. Overall, the evidence supports OSCD as a promising formulation for learning-based oil-spill detection, particularly when short temporal stacks of SAR imagery are available.

Taken together, these findings suggest that temporal supervision from synthetic pre-spill imagery is a powerful ingredient for robust localisation of oil-induced changes, but that it must be paired with architectures capable of effectively fusing and contrasting multi-temporal features. In our experiments, modules such as the cross-feature interaction blocks in DDLNet~\cite{ma2024ddlnetboostingremotesensing}, multi-scale fusion strategies, and attention-like mechanisms that highlight structures consistent across time emerge as key factors in suppressing residual artefacts while preserving fine-scale spill details. This indicates that further performance gains are likely to come from architectures that more explicitly model spatio-temporal context—such as fusion-aware transformers or joint CD–segmentation networks—rather than from exclusively refining single-frame segmenters alone.

\section{Conclusion}
We release the first bi-temporal OSCD dataset and present Temporal-Aware Hybrid Inpainting, a framework comprising two main components: High-Fidelity Hybrid Inpainting to restore oil-free backgrounds and Temporal Realism Enhancement to inject natural variability such as speckle and wave drift. Compared with traditional differencing and single-image segmentation, CD models trained on OSCD achieve significantly higher accuracy, confirming the value of temporally guided supervision in reducing false positives caused by look-alike phenomena.

More broadly, our work aligns with recent efforts to explicitly study temporal robustness and dataset evolution in geospatial vision, such as the CVTemporal benchmark for geo-localisation \cite{Deuser_2025_WACV}. OSCD plays a similar role for spill monitoring, offering a first step towards systematically assessing how CD models behave under realistic temporal variability and providing a basis for future multi-temporal extensions. From an application standpoint, OSCD-trained CD models can naturally be integrated into maritime surveillance chains as a temporal filtering stage that complements single-image detectors and expert analysis: candidate spills in a new SAR scene are cross-checked against a short temporal stack of recent acquisitions, allowing persistent look-alikes and artefacts to be down-weighted while emphasising genuinely new anomalies. The same idea can extend beyond oil pollution to other rare marine events wherever annotated post-event masks are available, with TAHI serving as a general tool for synthesising bi-temporal training data and OSCD as a template for constructing task-specific change detection benchmarks.

\bibliographystyle{IEEEtran}
\bibliography{main}
\end{document}